\documentclass[conference]{IEEEtran}
\usepackage{graphicx}
\usepackage{booktabs}
\usepackage{titlesec}
\usepackage{hyperref}
\usepackage{wrapfig}
\usepackage{devanagari}
\usepackage{multirow}

\begin{document}

\title{\LARGE Evaluating Machine Translation Models for English-Hindi Language Pairs: A Comparative Analysis}

\author{\authorblockN{\textbf {Ahan Prasannakumar Shetty}\authorrefmark{1}}
\authorblockA{\authorrefmark{1}National Institute of Technology Karnataka}}


\maketitle

\begin{abstract}
Machine translation has become a critical tool in bridging linguistic gaps, especially between languages as diverse as English and Hindi. This paper comprehensively evaluates various machine translation models for translating between English and Hindi. We assess the performance of these models using a diverse set of automatic evaluation metrics, both lexical and machine learning-based metrics.  Our evaluation leverages an 18000+ corpus of English-Hindi parallel dataset and a custom FAQ dataset comprising questions from government websites. The study aims to provide insights into the effectiveness of different machine translation approaches in handling both general and specialized language domains. Results indicate varying performance levels across different metrics, highlighting strengths and areas for improvement in current translation systems. 
\end{abstract}
\IEEEoverridecommandlockouts

\IEEEpeerreviewmaketitle

\section{Introduction}
Machine translation (MT) is pivotal in easing communication across linguistic boundaries, particularly between languages as distinct as English and Hindi. With the growth of digital content and global interaction, the demand for accurate and efficient translation systems has intensified, especially for low-resource languages. Evaluating the performance of machine translation models in such contexts is crucial for understanding their capabilities and limitations.\\
This paper comprehensively evaluates various machine translation models specifically designed for English-Hindi language pairs. The evaluation methodology encompasses a range of automatic evaluation metrics, including both lexical and machine learning-based approaches, to provide a nuanced assessment of translation quality and the quality deterioration as the sentence gets longer. Central to our study is the utilization of an extensive 18,000+ sentence parallel corpus of English-Hindi translations, which serves as a foundational dataset for benchmarking translation accuracy.\\
In an era marked by rapid digitization, governments worldwide are increasingly transitioning their paperwork and forms into digital formats, aiming to enhance accessibility and streamline services for their citizens. While promising greater efficiency and convenience, this transformation poses significant linguistic challenges, particularly for non-English speakers who encounter barriers in accessing and understanding critical information. This is most evident, especially in countries like India, which has a diverse linguistic population.
\\
For non-native English speakers, navigating through digital platforms for essential services such as healthcare, education, and legal documentation often requires proficient language skills, which many individuals may not possess. This linguistic gap underscores the critical need for robust machine translation (MT) systems capable of accurately converting information between languages, thereby enabling equitable access to government services and resources. The incorporation of an FAQ dataset, which spans questions regarding banking and tax regulation to public entrance exams, extends the evaluation scope to such domain-specific nuances.
\\
Through this comprehensive evaluation, this paper aims to contribute valuable insights into employing machine translation in areas where it could be deployed to improve information accessibility and bridge linguistic barriers.

\section{Evaluation Metrics}
Evaluating machine translation models is primarily aided by various evaluation metrics, which help quantify the quality of translations. Language variability results in lexical differences between translations, indicating no single correct translation. Evaluation done by a human is considered the golden standard due to their ability to capture semantic features. However, human evaluation comes with its own set of problems, such as inter-annotator agreement and inherent human bias, which is unavoidable. Human evaluation is also a time and resource exhaustive process. \\
Automatic evaluation metrics, on the other hand, are a more cost- and time-effective approach to evaluating translations. However, their quality isn't on par with human evaluation. \\
Based on the approach taken by these metrics, automatic evaluation can be classified into the following categories \cite{cantrell1}: 
\begin{itemize}
  \item {Lexical-based metrics}
  \item {Machine Learning- based metrics}

\end{itemize}
\begin{table*}[h]
  \caption{Example of problems with automatic evaluation metrics. Similarity refers to the human-annotated semantic similarity between sentence 1 and sentence 2. \cite{cantrell2}}
  \centering
  \begin{tabular*}{\textwidth}{@{\extracolsep{\fill}} p{4cm} p{4cm} c c c @{}}
    \toprule
    \textbf{Sentence 1} & \textbf{Sentence 2} & \textbf{Similarity} & \textbf{BLEU} & \textbf{BLEURT} \\
    \midrule
    An arrest warrant claimed Bryant\newline assaulted the woman 30 June at a hotel. & According to an arrest warrant, Bryant, 24, attacked a woman on 30 June. & 85.0 & 8.72 & 71.28 \\
    \midrule
    Two white dogs are swimming in the water. & The birds are swimming in the water. & 16.0 & 66.42 & 43.38 \\
    \bottomrule
  \end{tabular*}
  \label{tab:problems}
\end{table*}

\subsection{Lexical-based Metrics}
These have been used extensively to evaluate machine translations due to their simplicity and ease of computing. Their evaluation basis is measuring the overlap between human-annotated and machine translations. Since they are simple in architecture, they do not require pre-training and can be directly employed to churn out a score.\\ However, a major drawback of such a system is its inability to capture semantic meaning. A given word may have multiple acceptable translations that cannot be evaluated by mere overlapping.\\
As shown in Table \ref{tab:problems}, automatic evaluation metrics can have limitations. Although the first example of sentences gives the same context,  BLEU (a popular lexical-based automatic evaluation metric) gives it a poor score. On the other hand, the sentences in the second example are completely unrelated, but BLEU gives them a higher score because of their high overlap.\\
Following are the lexical metrics used in our evaluation \cite{cantrell2}:

\subsubsection{BLEU:}

Bilingual Evaluation Understudy (BLEU) is a word-based automatic evaluation metric. It uses n-gram-based precision. This metric counts the frequency of n-gram matches between hypothesis and reference and uses a brevity penalty to penalize predictions that are too short. It is a widely used metric.

\subsubsection{WER:}
Word error rate is the easiest metric to calculate. It measures the edit distance between the hypothesis and reference in terms of measurable insertions, deletions, and substitutions to achieve the reference from the hypothesis.

\subsubsection{TER:}
 Translation Error Rate (TER) is similar to WER but adds an extra editing step, which allows for shifting. TER only focuses on word-level matching and not fluency.

 \subsection{Machine Learning-based Metrics}
 These metrics capture the similarity between hypothesis and reference using word/contextual embedding models. This shows a much higher correlation with human judgment because of its ability to capture semantic meaning. Machine Learning-based metrics require models to be trained on data with source and target languages and may show poor correlation in the case of low-resource languages. Following are the machine learning-based metrics used in our evaluation\cite{cantrell2}: 

 \subsubsection{BLEURT:}
 Bilingual Evaluation Understudy with Representations from Transformers \cite{cantrell2} is a pre-trained model using the BERT structure. It is a sentence-level metric that learns to predict scores for the similarity between hypothesis and reference. 

 \subsubsection{BERTScore:}
BERTScore \cite{cantrell3} compares the hypothesis and reference by using the features extracted by BERT model. It uses individual word embeddings and surrounding words to provide a better understanding of the context. This evaluation metric computes the cosine similarity between the tokenized embeddings of the hypothesis and the reference.

 \subsubsection{COMET:}
 Cross-lingual optimized metric for evaluation of translation  (COMET) \cite{cantrell4} is a metric for multilingual translation evaluation using ranking and regression. The use of dual cross-lingual modeling shows improved performance since it also considers the source text on top of the reference text to provide a better context during evaluation.

\section{Machine Translation Models}
Machine Translation (MT) models were selected based on their ease of deployment, support for English-Hindi translations, and accessibility, which made open-source models the perfect candidates for evaluation. An extensive literature survey on existing work narrowed the search down to a few commonly used MT models/APIs. The following models have been evaluated in this paper
\subsubsection{NLLB-200}
No Language Left Behind (NLLB) \cite{cantrell5} is an open-source machine translation (MT) model developed by Meta AI. The variant used for our evaluation is a 600M-distilled variant and is based on the transformer encoder-decoder architecture; however, every 4th Transformer block has its feed-forward layer replaced with a Sparsely Gated Mixture of Experts layer. With support for translations across 200 languages and a robust yet efficient neural architecture, it is a strong candidate for deployment in real-world applications.

\subsubsection{Google Translate}
One of the most widely used translation services for day-to-day tasks, Google Translate, is known for accurate translations and easy-to-use interface. It uses an example-based machine translation (EBMT) method wherein the system learns from millions of examples and uses a broader context to figure out the most relevant translation. With a reliable and cost-effective API service, Google Translate can easily be deployed in various applications.

\subsubsection{OPUS-MT}
Based on the Marian-NMT framework, OPUS-MT \cite{cantrell6} offers both bilingual and multilingual models. With over 1000 pre-trained translation models that are open-source, it facilitates easy deployment and integration. The architecture is based on a standard transformer setup with 6 self-attentive layers in both the encoder and decoder network with 8 attention heads in each layer. 

\subsubsection{IndicTrans2}
One of the only open-source translation models supporting translations across all 22 scheduled languages of India, IndicTrans2 \cite{cantrell7} developed by AI4Bharat focuses on the unique linguistic characteristics of Indian languages. Trained on a large and extensive dataset, it is based on a transformer encoder-decoder architecture that aims at improving translation quality and handling nuances and idiomatic expressions specific to Indian languages.

\section{Evaluation Methodology}
\subsection{Corpora}
The corpora used are described in the \href{https://github.com/ahanps/English-Hindi-Parallel-Corpus}{Corpus Repository}\footnote{See the corpus repository at \url{https://github.com/ahanps/English-Hindi-Parallel-Corpus}}.
The assessment was performed using two datasets: an extensive set of over 18,000 parallel English-Hindi sentence pairings and a smaller, more specialized FAQ dataset comprising approximately 400 English-Hindi parallel Question-Answer pairings.\\
\subsubsection{General Parallel Corpus}
This dataset consists of over 18,000 parallel sentence pairs in English and Hindi. The sentences in this in-house corpus span a wide range of topics and contexts, providing a comprehensive foundation for assessing the overall performance of the MT models.
\subsubsection{FAQ Corpus}
This specialized dataset consists of approximately 400 English-Hindi parallel question-answer pairings. The included FAQs have been sourced from frequently accessed online platforms and encompass a wide range of subjects, including government forms and banking-related queries. The dataset was meticulously curated to assess the efficacy of models in translating domain-specific content, a fundamental requirement for practical applications.

\subsection{Unidirectional Translation}
The models were utilized for the translation of sentences, followed by a comparison of the translated sentences with the respective reference sentences. Furthermore, the assessment involved calculating the variability of scores in relation to varying sentence lengths, thereby providing insight into the model's capacity to comprehend context amidst increasing text length. None of the metrics or MT models underwent any pre-training and have been evaluated on the current (as of writing this paper) publicly available stock versions.

\subsection{Back-Translation}
In addition to unidirectional translation, the technique of back-translation has been incorporated. This involves translating sentences from the source language to a target language and then re-translating them back to the source language. This process serves to validate the faithfulness of the translation. A close alignment between the back-translated sentence and the original indicates that the model has effectively preserved the semantic and contextual aspects during translation. Assessment of the model's proficiency in handling back-translation enables the evaluation of its robustness. A robust model should adeptly translate text between languages without substantial loss of information or introduction of errors. This is particularly crucial in ensuring the model's capability to navigate intricate and nuanced sentences without compromising quality \cite{cantrell8}.\\
Moreover, back-translation can reveal subtle semantic and contextual errors that may not be apparent in unidirectional translation.

\section{Results}
\subsection{Unidirectional Translation}
Following are the results depicted for the unidirectional translations of the general 18,000+ sentence pairings corpus. \smallbreak
\begin{figure}[h!]
  \centering
  \includegraphics[width=0.475\textwidth]{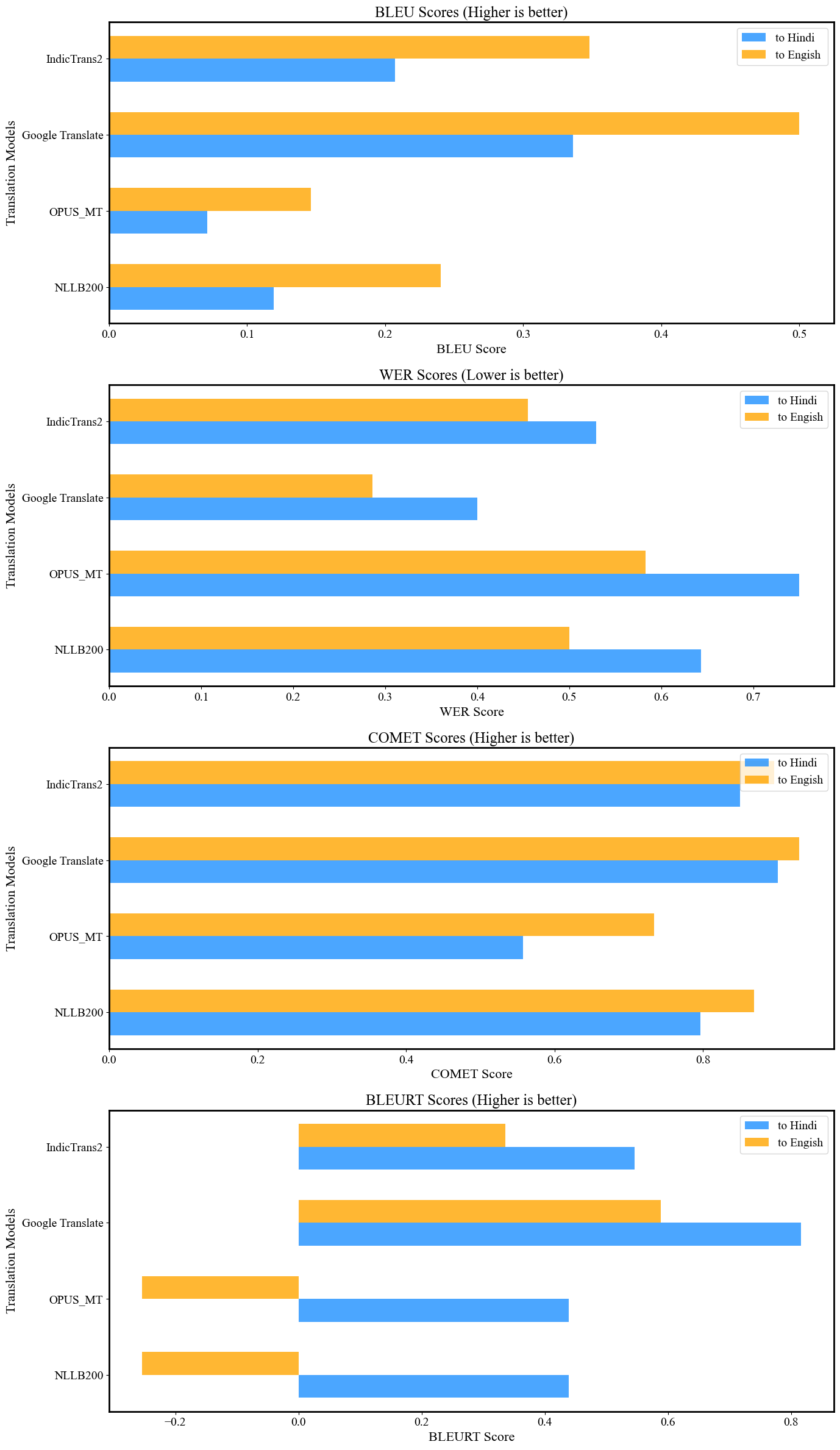}
  \caption{MT Model median scores for Unidirectional Translation (on 18,000+ sentence pair corpus)}
  \label{fig:my_plot}
\end{figure}
In the visualization depicted in Figure \ref{fig:my_plot}, it is apparent that Google Translate consistently delivers superior performance when compared to other models, closely followed by IndicTrans2. BLEURT scores can be negative because they reflect ratings from the WMT Metrics Shared Task, where scores are normalized per annotator. A negative score indicates that the quality of the generated text is significantly lower than average when compared to the reference data. In contrast, a positive score suggests a higher quality alignment. Specifically, OPUS MT and NLLB200 exhibit negative scores, indicating below-average translations, especially when translating from a resource-scarce language like Hindi.  Furthermore, the machine translation models demonstrate significantly better results when translating from Hindi to English, as opposed to the reverse direction. Additionally, Google Translate exhibits more consistent performance, as reflected in its lower standard deviation values in comparison to the deviations of the other models from their mean scores.

\begin{figure}[h!]
  \centering
  \includegraphics[width=0.475\textwidth]{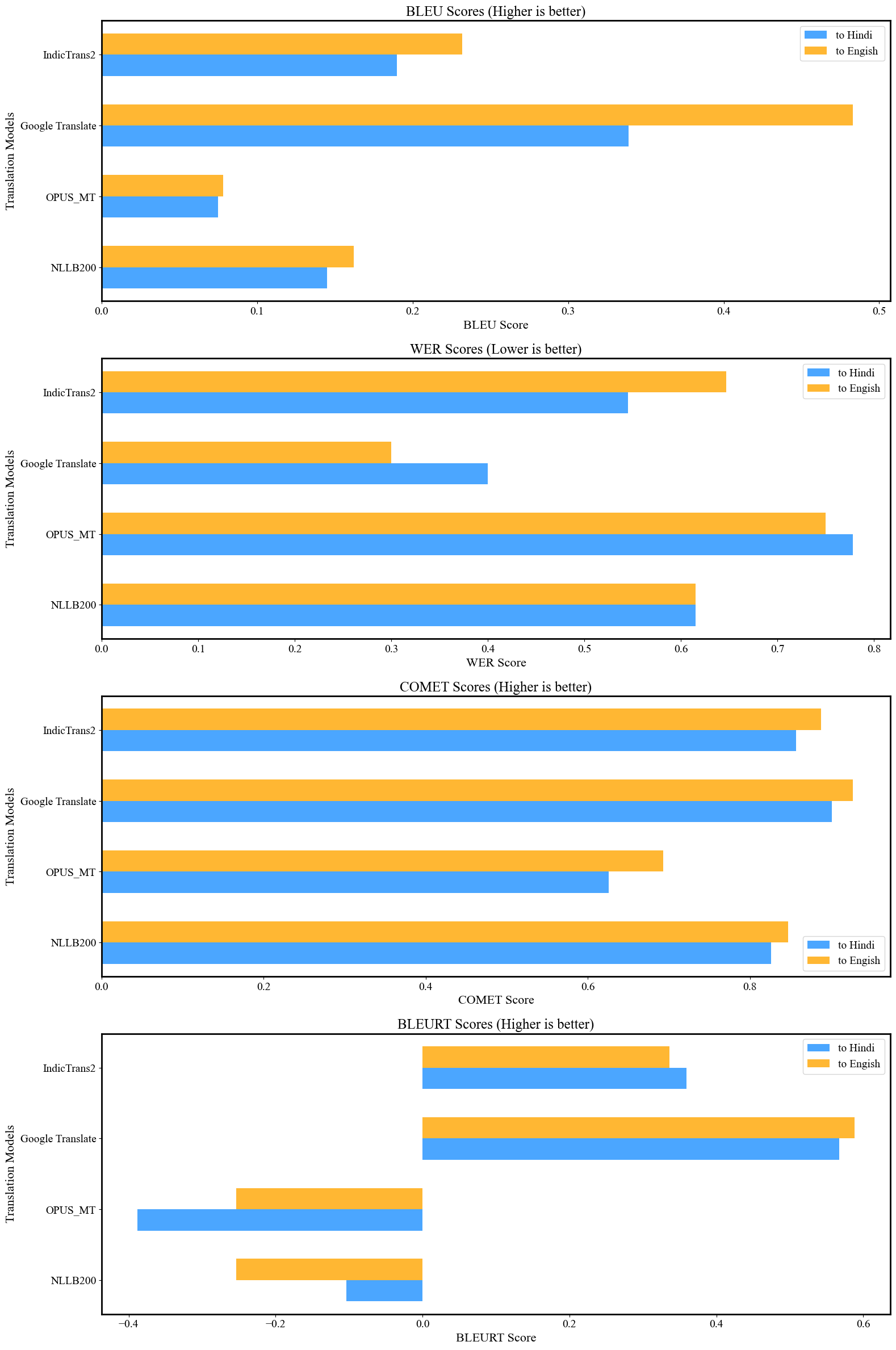}
  \caption{MT Model median scores for Back-Translation (on 18,000+ sentence pair corpus)}
  \label{fig:my_plot2}
\end{figure}

\subsection{Effect of Word Count on Translation Performance}

The translation quality of all evaluated models is significantly impacted by the increase in word count.
All models show degrading translation accuracy as the word count in sentences being translated increases.\\
One of the primary factors contributing to this observation lies in the inherent difficulty of training machine translation models using lengthy sentences. This challenge stems from the limited availability of training corpora containing an adequate number of extended sentences and the linear correlation between the computational overhead of each training update iteration and the length of the training sentences \cite{cantrell9}.

Furthermore, the neural network may fail to encapsulate all details due to the method of encoding a variable-length sentence into a fixed-size vector representation, commonly employed by most models.

\begin{figure}[h!]
  \centering
  \includegraphics[width=0.45\textwidth]{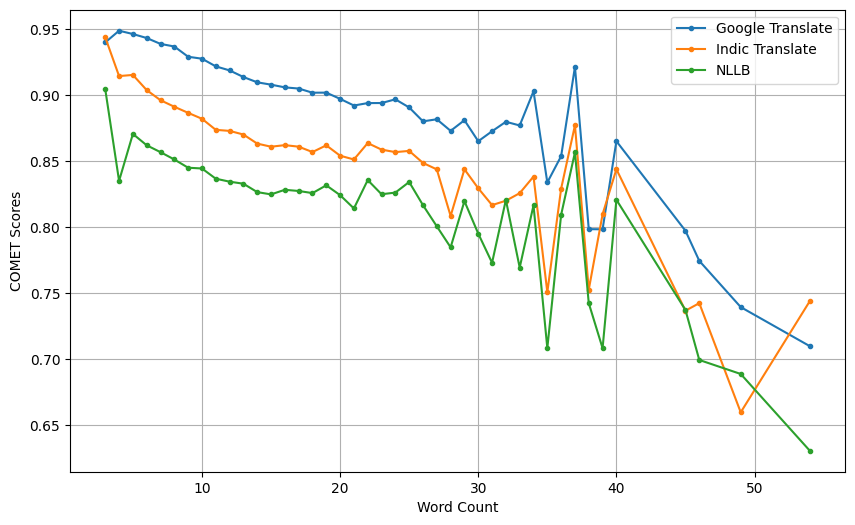}
  \caption{COMET Score vs Word Count}
  \label{fig:my_plot3}
\end{figure}

\subsection{Evaluating Poor Translations}
\subsubsection{Gender Marking}
English is largely gender-neutral in its structure, with the exception of pronouns (he, she, they) and some gender-specific nouns. Hindi, on the other hand, is widely a gendered language where nouns, adjectives, and verbs all carry gender markers. One of the main failing points of machine translation models is their inability at times to properly interpret these gender markers whilst translating.
\begin{figure}[h!]
  \centering
  \includegraphics[width=0.45\textwidth]{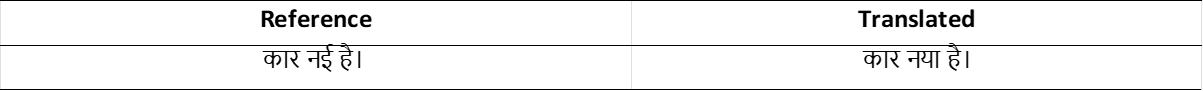}
  \label{fig:my_plot5}
\end{figure}
The word {\dn kAr} (work) is a feminine noun; thus, the adjective describing it should be feminine.
\\There is also the issue of gender bias in training data. When the training corpus over-represents specific gendered terms or roles, it can lead to the development of a biased perspective in the machine translation (MT) model. For instance, if the data primarily includes male doctors and female nurses, the model may inaccurately assign these genders during translation.
\begin{figure}[h!]
  \centering
  \includegraphics[width=0.35\textwidth]{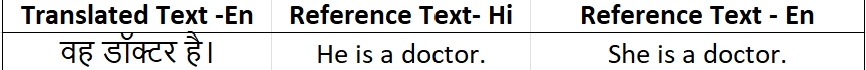}
  \label{fig:my_plot5}
\end{figure}

\begin{table*}[ht]
  \caption{Unidirectional Translation Scores on FAQ Corpus}
  \centering
  \begin{tabular*}{\textwidth}{@{\extracolsep{\fill}} l | l | ccc | ccc | ccc @{}}
    \toprule
    \textbf{Model\hspace{20pt}} & \textbf{Type} & \multicolumn{3}{c}{\textbf{BLEU}} & \multicolumn{3}{c}{\textbf{WER}} & \multicolumn{3}{c}{\textbf{TER}} \\
    \midrule
    & & \textbf{Mean} & \textbf{Median} & \textbf{Std Dev} & \textbf{Mean} & \textbf{Median} & \textbf{Std Dev} & \textbf{Mean} & \textbf{Median} & \textbf{Std Dev} \\
    \midrule
    \multirow{2}{*}{NLLB-200} 
    & Qn: En to Hi & 0.366 & 0.302 & 0.208 &  0.506& 0.5 & 0.276 & 0.471 & 0.45 & 0.249 \\
    & Qn: Hi to En & 0.339 &0.274  & 0.189 & 0.542 & 0.5 & 0.297 & 0.514 & 0.5 & 0.272 \\
    & Ans: En to Hi & 0.3 & 0.261 & 0.16 & 0.629 & 0.637 &  0.246& 0.57 & 0.588 & 0.232 \\
    & Ans: Hi to En & 0.265 & 0.239 & 0.138 & 0.654 & 0.634 & 0.263 & 0.605 & 0.577 & 0.249 \\
    \midrule
    \multirow{2}{*}{Google Translate} 
    & Qn: En to Hi & 0.422 & 0.357 & 0.233 & 0.413 & 0.4 & 0.252 & 0.39 & 0.381 &  0.236\\
    & Qn: Hi to En & 0.458 & 0.38 & 0.252 & 0.389 & 0.364 & 0.278 & 0.373 & 0.333 & 0.264 \\
    & Ans: En to Hi & 0.366 & 0.329 & 0.187 & 0.546 & 0.544 & 0.254 &0.494  & 0.5 & 0.242 \\
    & Ans: Hi to En & 0.412 & 0.394 &0.191  & 0.479 &0.455  & 0.235 & 0.437 &0.423  & 0.214 \\
    \midrule
    \multirow{2}{*}{IndicTrans2} 
    & Qn: En to Hi & 0.403 & 0.316 & 0.238 & 0.492 & 0.455 & 0.321 & 0.471 & 0.444 & 0.311 \\
    & Qn: Hi to En & 0.41 & 0.332 &0.227  & 0.466 & 0.444 & 0.3 & 0.444 &0.429  &0.28  \\
    & Ans: En to Hi & 0.341 & 0.294 & 0.19 & 0.607 & 0.633 & 0.282 &0.552  & 0.55 & 0.269 \\
    & Ans: Hi to En & 0.349 & 0.324 & 0.17 & 0.573 & 0.549 & 0.252 & 0.526 & 0.524 &0.233  \\
    \bottomrule
  \end{tabular*}
  \label{tab:comparison}
\end{table*}

\begin{table*}[ht]
  \centering
  \begin{tabular*}{\textwidth}{@{\extracolsep{\fill}} l | l | ccc | ccc | ccc @{}}
    \toprule
    \textbf{Model \hspace{20pt}} & \textbf{Type} & \multicolumn{3}{c}{\textbf{COMET}} & \multicolumn{3}{c}{\textbf{BLEURT}} & \multicolumn{3}{c}{\textbf{BERTScore}} \\
    \midrule
    & & \textbf{Mean} & \textbf{Median} & \textbf{Std Dev} & \textbf{Mean} & \textbf{Median} & \textbf{Std Dev} & \textbf{Mean} & \textbf{Median} & \textbf{Std Dev} \\
    \midrule
    \multirow{2}{*}{NLLB-200}
    & Qn: En to Hi & 0.865 & 0.883 & 0.078 & 0.664 & 0.709 & 0.242 & 0.907 & 0.908 &0.052  \\
    & Qn: Hi to En & 0.904 & 0.912 & 0.058 & 0.402 & 0.432 & 0.324 & 0.849 & 0.849 & 0.082 \\
    & Ans: En to Hi & 0.791 & 0.797 & 0.077 & 0.503 & 0.489 &  0.16&  0.889 &0.891 &0.043 \\
    & Ans: Hi to En & 0.847 & 0.855 & 0.065 & 0.233 & 0.279 & 0.313 & 0.821 &  0.826& 0.076 \\
    \midrule
    \multirow{2}{*}{Google Translate} 
    & Qn: En to Hi & 0.893 & 0.906 & 0.065 & 0.738 & 0.829 & 0.224 & 0.924 & 0.929 & 0.051 \\
    & Qn: Hi to En & 0.933 & 0.942 & 0.041 & 0.609 & 0.642 & 0.272 & 0.898 & 0.904 & 0.079 \\
    & Ans: En to Hi & 0.827 & 0.828 & 0.054 & 0.528 & 0.497 & 0.139 & 0.903 & 0.904 &0.038  \\
    & Ans: Hi to En & 0.884 & 0.886 & 0.041 & 0.456 & 0.469 & 0.241 & 0.884 & 0.891 & 0.069 \\
    \midrule
    \multirow{2}{*}{IndicTrans2} 
    & Qn: En to Hi & 0.874 & 0.887 & 0.075 & 0.726 & 0.798 & 0.207 & 0.911 & 0.915 & 0.056 \\
    & Qn: Hi to En & 0.926 & 0.935 & 0.044 & 0.564 & 0.598 & 0.283 & 0.887 & 0.895 & 0.081 \\
    & Ans: En to Hi & 0.804 & 0.81 & 0.07 & 0.518 & 0.491 & 0.157 & 0.892 & 0.894 & 0.046 \\
    & Ans: Hi to En & 0.87 & 0.874 & 0.054 & 0.381 & 0.413 & 0.286 & 0.861 &  0.869& 0.073 \\
    \bottomrule
  \end{tabular*}
  \label{tab:comparison}
\end{table*}

\subsubsection{Abbreviations}
Machine translation (MT) models often encounter difficulties in accurately translating English abbreviations, frequently resulting in untranslated or preserved abbreviations within the target text. 
\begin{figure}[h!]
  \centering
  \includegraphics[width=0.45\textwidth]{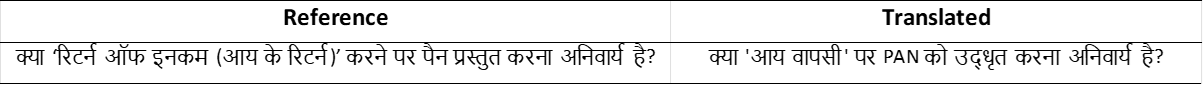}
  \label{fig:my_plot6}
\end{figure}
The term PAN (Permanent Account Number) is left untranslated, creating a substantial barrier for those who are not proficient in the English alphabet. This poses a serious challenge for many individuals, hindering their ability to navigate systems in real life effectively.
The significance of this becomes particularly pronounced in the realm of domain-specific translations, such as those pertaining to banking and government forms FAQs, which frequently encompass numerous abbreviations.

\subsubsection{Proverbs and Common Sayings}
Proverbs are highly cultural and region-specific, making it challenging to translate them accurately. MT models often fail to capture the idiomatic and cultural nuances of sayings, leading to confusion or loss of meaning.
\begin{figure}[h!]
  \centering
  \includegraphics[width=0.45\textwidth]{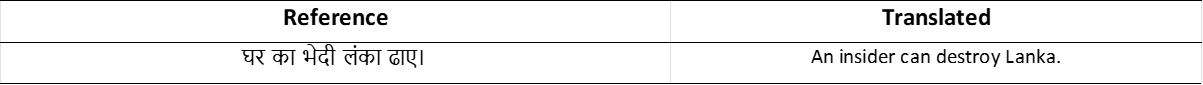}
  \label{fig:my_plot7}
\end{figure}

The literal translation in the above case is not clear in English. The correct translation conveys the idea that someone who knows the inner workings can cause significant damage.

\subsubsection{Lost Semantic Meaning}
Certain words pertain to different meanings based on the context in which they are used. However, MT models regularly failed to recognize the contextual use of the words and would often churn out a translation based on either the literal meaning or in the context most commonly used.
\begin{figure}[h!]
  \centering
  \includegraphics[width=0.45\textwidth]{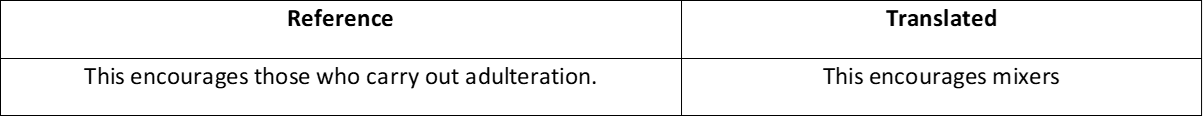}
  \label{fig:my_plot8}
\end{figure}

\vfil\eject

\section{Conclusion}
The findings underscore the robustness of Google Translate, which can be attributed to the extensive dataset accessible to the system. This large dataset enables Google Translate to leverage context extraction capabilities, a critical feature in machine translation. Context extraction refers to the system's ability to analyze and interpret the surrounding text (context) to resolve ambiguities, maintain consistency, and capture nuanced meanings. For example, it can determine whether the word "bank" refers to a financial institution or the side of a river based on the surrounding words and sentences. This capability allows Google Translate to produce translations that are not only accurate but also coherent and contextually appropriate, especially in longer or more complex texts.
\\
IndicTrans2 performs competitively, consistently ranking second, and offers the added benefit of supporting translation across all officially recognized languages of India. However, during the evaluation of back-translation—a process that checks for truthfulness (accuracy of translation) and consistency (logical flow and alignment with the source text)—Google Translate demonstrated superior performance. This is largely due to its advanced context extraction capabilities, which enable it to better interpret and preserve the meaning, tone, and intent of the original text \cite{cantrell16}. In contrast, while IndicTrans2 is effective, it may not match Google Translate's ability to handle highly nuanced or context-dependent translations, particularly in scenarios requiring deep contextual understanding.


\vfil\eject

\smallskip

\end{document}